\documentclass[journal]{IEEEtran}
\IEEEoverridecommandlockouts
\usepackage{times}  
\usepackage{graphicx}
\usepackage{amsmath,amssymb,amsfonts}
\usepackage{cite}
\usepackage{hyperref}
\usepackage{booktabs}
\usepackage{caption}
\usepackage{subfig} 
\usepackage{float}
\IEEEpubid{\makebox[\columnwidth]{979-8-3315-2261-2/24/$31.00~\copyright2025 IEEE \hfill}\hspace{\columnsep}\makebox[\columnwidth]{ }}
\begin{document}

\title{Continual Learning for Food Category Classification Dataset: Enhancing Model Adaptability and Performance}

\author{
\begin{minipage}[t]{0.3\textwidth}
\centering
\IEEEauthorblockN{Piyush Kaushik Bhattacharyya} \\
\IEEEauthorblockA{\textit{School of Computer Engineering} \\
\textit{KIIT University} \\
Bhubaneshwar, India \\
piyushbhattacharyya@gmail.com}
\end{minipage}\hfill
\begin{minipage}[t]{0.3\textwidth}
\centering
\IEEEauthorblockN{Devansh Tomar} \\
\IEEEauthorblockA{\textit{School of Computer Engineering} \\
\textit{KIIT University} \\
Bhubaneshwar, India \\
devanshtomar100@gmail.com}
\end{minipage}\hfill
\begin{minipage}[t]{0.3\textwidth}
\centering
\IEEEauthorblockN{Shubham Mishra} \\
\IEEEauthorblockA{\textit{School of Computer Engineering} \\
\textit{KIIT University} \\
Bhubaneshwar, India \\
shubhamlala63@gmail.com}
\end{minipage}

\vspace{1em}

\begin{minipage}[t]{0.3\textwidth}
\centering
\IEEEauthorblockN{Divyanshu Rai} \\
\IEEEauthorblockA{\textit{School of Computer Engineering} \\
\textit{KIIT University} \\
Bhubaneshwar, India \\
raidivyanshu3232@gmail.com}
\end{minipage}\hfill
\begin{minipage}[t]{0.3\textwidth}
\centering
\IEEEauthorblockN{Yug Pratap Singh} \\
\IEEEauthorblockA{\textit{School of Computer Engineering} \\
\textit{KIIT University} \\
Bhubaneshwar, India \\
yugpratap28@gmail.com}
\end{minipage}\hfill
\begin{minipage}[t]{0.3\textwidth}
\centering
\IEEEauthorblockN{Harsh Yadav} \\
\IEEEauthorblockA{\textit{School of Computer Engineering} \\
\textit{KIIT University} \\
Bhubaneshwar, India \\
hy2002vk18@gmail.com}
\end{minipage}

\vspace{1em}

\begin{minipage}[t]{0.3\textwidth}
\centering
\IEEEauthorblockN{Krutika Verma} \\
\IEEEauthorblockA{\textit{School of Computer Engineering} \\
\textit{KIIT University} \\
Bhubaneshwar, India \\
krutika.vermafcs@kiit.ac.in}
\end{minipage}\hfill
\begin{minipage}[t]{0.3\textwidth}
\centering
\IEEEauthorblockN{Vishal Meena} \\
\IEEEauthorblockA{\textit{School of Computer Engineering} \\
\textit{KIIT University} \\
Bhubaneshwar, India \\
vishal.meenafcs@kiit.ac.in}
\end{minipage}\hfill
\begin{minipage}[t]{0.3\textwidth}
\centering
\IEEEauthorblockN{N Sangita Achary} \\
\IEEEauthorblockA{\textit{School of Computer Engineering} \\
\textit{KIIT University} \\
Bhubaneshwar, India \\
sangita.acharyfcs@kiit.ac.in}
\end{minipage}
}

\maketitle
\IEEEpubid{\makebox[\columnwidth]{979-8-3315-1375-7/25/\$31.00~\copyright~2025 IEEE \hfill}
\hspace{\columnsep}\makebox[\columnwidth]{ }}

\IEEEtitleabstractindextext{
\begin{abstract}
Conventional machine learning pipelines often struggle to recognize categories absent from the original training set. This gap typically reduces accuracy, as fixed datasets rarely capture the full diversity of a domain. To address this, we propose a continual learning framework for text-guided food classification\cite{b1, b2}. Unlike approaches that require retraining from scratch, our method enables incremental updates, allowing new categories to be integrated without degrading prior knowledge\cite{b3, b5, b6}. For example, a model trained on Western cuisines could later learn to classify dishes such as dosa or kimchi\cite{b13, b14}. Although further refinements are needed, this design shows promise for adaptive food recognition, with applications in dietary monitoring and personalized nutrition planning\cite{b15, b10}.
\end{abstract}

}

\IEEEdisplaynontitleabstractindextext

\section{Introduction}
Food recognition has gained importance due to its role in assisting diet tracking, providing personal nutritional information and recommending menus \cite{b1}. On the other hand, legacy machine learning is not without its share of shortcomings. They rely on static datasets and hence, are unable to discover new foods or cope with adaptive eating behaviours. Case in point: a model trained primarily on Western food can never pick up regionals like Kimchi or dosa. Thus, the multiculturally originated and dynamic nature of the data on foods necessitates continuous updates in an effort to maintain the quality of the model \cite{b2}.\\
\noindent
 As steps towards addressing these, incrementally learned methods have been proposed as a means for giving models a greater ability to adapt to newer information while maintaining previously learned knowledge \cite{b3,b4}. They are promising yet they are not without issues. Catastrophic forgetting, where previously learned knowledge is lost while learning new data, remains a ubiquitous problem \cite{b5,b6}. Some mitigation methods, for example, gradient episodic memory \cite{b6} and knowledge distillation \cite{b7}, have been proposed. They are only patch solutions and may be affected by task and dataset complexity and shift across domains.
\noindent
The key contributions of this study are as follows:
\begin{itemize}
    \item Initial application of continual learning in the domain of food classification
    \item Robust Evaluation and Enhanced Generalization
    \item Minimized Human Intervention and Scalable Learning
\end{itemize}
\noindent
\section{Related Works}
\begin{figure}[!htbp]
    \centering
    \includegraphics[width=0.5\textwidth]{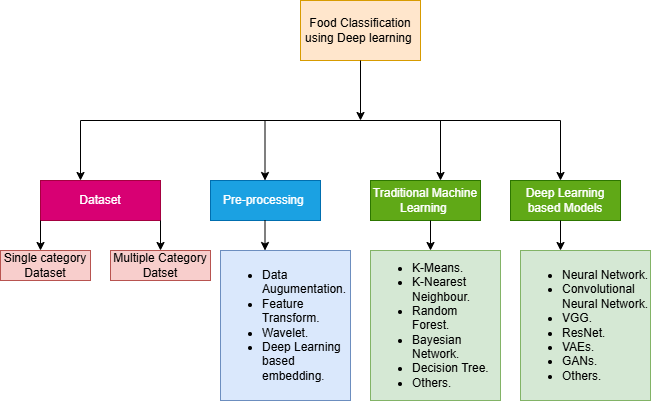}
    \caption{\textbf{Taxonomy of Food Classification using Deep Learning.}}
    \label{avg}
\end{figure} 
\noindent
 Recent developments in the classification of foods have been influenced mainly by the availability of large and diverse datasets, which have allowed for rapid progress in the formulation of recognition techniques based on deep learning. Among them, Food-101 has gained wide acceptance as a benchmark, comprising 101,000 images distributed across 101 different categories of foods~\cite{b13}. Similarly, UECFOOD-256 has gained wide use as a set for Japanese foods, offering 256 categories complete with bounding box information, thereby allowing for classification as well as precise localization tasks~\cite{b14}. VireoFood-172 offers another valuable addition, particularly in a multimodal context, by correlating visual information with lists of constituents and thereby allowing for more elaborate cross-modal explorations~\cite{b15}. As a group, these datasets have influenced the course of research, lying at the foundation of innovation in image-driven recognition for foods, detection at the level of ingredients, and the conjoining of visual and textual modalities within a generic learning paradigm. For example, while a model formulated for learning on Food-101 may be suitable for generic recognition of dishes, it may be inadequate in correctly discerning region specialties such as Japanese okonomiyaki or Korean bibimbap unless supplemented by more specific datasets such as UECFOOD-256 or VireoFood-172.\\
\noindent
 Progressive Prompt approach provides a computationally light and conceptually easy model for continuous learning in language models. The approach supports one-directional forward knowledge transfers while simultaneously restricting the chance for catastrophic forgetting without data replay or a plethora of task-specialized parameters. The approach as such involves the learning of a soft prompt for each task that is then incrementally appended to the history of previously encountered prompts. The approach augments the capacity for a model without a modification to the underlying base architecture\cite{16}.\\
\noindent
 Convolutional Neural Networks (CNNs) are the current go-to approach for classifying food images. Deep learning models have been found to perform exceptionally well given large-scale datasets in the literature \cite{b1}. The problem yet to be addressed is that they perform poorly in generalizing to hitherto unknown food categories \cite{b2}. In a bid to compensate for that weakness, transfer learning methods are increasingly being adopted, which involve fine-tuning the pre-trained networks to accomplish a specified task. The drawback to such methods is that they quite often fail when class imbalance is involved \cite{b3}.\\
\noindent
As an approach to transcend the constraint inherent in fixed learning structures, CL has been a promising way to counter catastrophic forgetting. CL enables a model to retain past knowledge while improved accommodation of new observed classes of foods is incorporated into classification structures. Methods like Elastic Weight Consolidation (EWC) \cite{b6}, experience replay \cite{b7}, and knowledge distillation \cite{b8} have been rigorously researched for natural language processing settings. Nevertheless, in the context for the field of food image classification, they are comparatively and yet underutilized, with great promise to be harnessed in future research and development. 

\section{Methodology}
\subsection{Problem Statement}
\noindent
In this study, we address the adaptive categorization problem of foods and constructed a Continual Learning model that was designed with custom Indian meal inputs to distinguish between Veg and Non-Veg category. The high-level idea is to build a model that can accurately categorize a previously seen categories while able to understand and classify novel elements. Incremental learning is considered for inserting newly labeled data into the model that refrains from a whole retraining. The strategy ensures that the classification system is flexible and expandable and permits ongoing augmentation when more data are received.

\begin{figure*}[!htbp]
        \centering
        \includegraphics[width=0.4\textwidth]{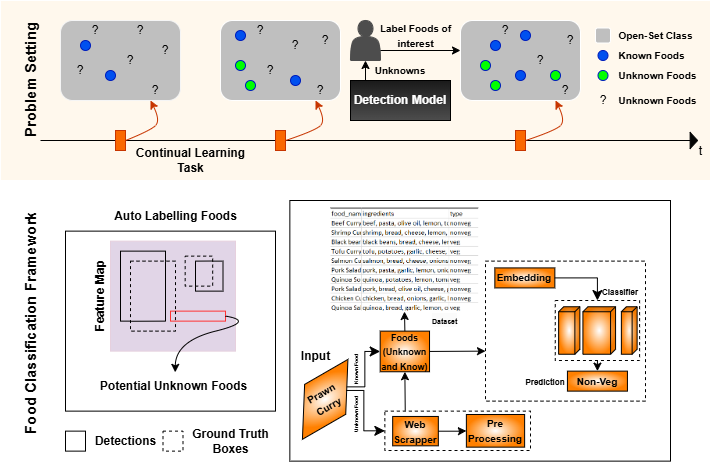}
        \label{fig:ao}
       \caption{\textbf{Proposed Framework}  Top Row: The model at each step in the incremental learning process is provided with novel previously unknown food items, denoted by a question mark (‘?’). As they get labeled (denoted by blue circles), they become incorporated into the model's current knowledge base without delay, denoted by green circles. This continuous learning scheme ensures that the model expands its capacity for understanding foods while at the same time maintaining the knowledge for already learned food categories.\\   
 \textit{Bottom Row:} The proposed Food Classification Framework would ideally be capable of differentiating and processing hitherto unheard-of foods. The first auto-labelling approach is borrowed for the identification of potential unknown food groups by scanning feature representations at the feature map level. The unknown entities are thereby mapped into the feature space, thereby distinctively separating them from known food groups. Given a hitherto unheard-of food item (e.g., "Prawn Curry"), a TF-IDF-based classification is conducted on the item such that the resultant features are fed under a classifier. The system makes the right prediction for the category of the food (e.g., "Non-Veg"), thereby ensuring that the integration of newly learned food categories doesn't hinder or dwindle the knowledge gained by hitherto observed classes.}

\end{figure*}
\subsection{Proposed Framework}
\noindent
 The feature map for the system we propose is the TF-IDF (Term Frequency–Inverse Document Frequency) vectors of the relevant textual data for foods, i.e., lists for dishes and ingredients. The TF-IDF representation captures well the relative term importance for the entire dataset and offers a semantically informative and interpretable representation for the textual input.

\subsection{Proposed Neural Network Architecture}
\noindent
 The approach applies a feedforward neural network for classification of dishes into binary categories \textit{Vegetarian} and \textit{Non-Vegetarian} from textual features. The input data are represented as Term Frequency–Inverse Document Frequency (TF-IDF) vectors calculated for the names of the dishes. Feature extraction is limited to a maximum of 5000 terms for a tradeoff between relevance and discriminability. The stop-words are eliminated.\\
\noindent
 The neural network coded in Keras is structured in a sequence of layers as shown below:

\begin{itemize}
 \item \textbf{Input Layer:} The model is initiated with an input layer that is given a Term Frequency-Inverse Document Frequency (TF-IDF) feature vector of dimension 5000, indicating the importance of terms in the entire corpus.
 \item \textbf{First Hidden Layer:} The initial hidden layer is a dense (fully connected) one with 64 neurons, and the activation function that we employ is the Rectified Linear Unit (ReLU). As one type of defense against the threat of overfitting, we use L2 regularization with penalty coefficient 0.01.
 \item \textbf{Second Hidden Layer:} The first dense layer with 32 neurons comes next, and we use the same ReLU activation function and same value for the L2 regularization factor (0.01) for purposes of facilitating generalization while in training.
 \item \textbf{Output Layer:} The output layer of the network is a single sigmoid-activated neuron that permits binary classification by outputting a value between 0 and 1 as a probability.
\end{itemize}

A summary of the key hyperparameters adopted in this study is provided below:

\begin{table}[h]
\centering
\caption{Model Hyperparameters}
\resizebox{.3\textwidth}{!}{
\begin{tabular}{|l|l|}
\hline
\textbf{Parameter} & \textbf{Value} \\
\hline
TF-IDF max features & 5000 \\
Batch size & 32 \\
Epochs & 100 \\
Optimizer & Adam \\
Loss function & Binary Crossentropy \\
Hidden layer sizes & 64, 32 \\
Activation (hidden layers) & ReLU \\
Activation (output layer) & Sigmoid \\
Regularization & L2 (0.01) \\
Early stopping patience & 5 \\
\hline
\end{tabular}
}
\end{table}

\subsection{Data Collection}
\noindent
  The web scraping techniques were used to obtain semi-automatic and systematic lists of dish names from various sources on the Internet, for example, on Wikipedia and other reliable culinary sites. The step of data acquisition was accomplished with the use of Python's \textit{BeautifulSoup} package which enabled proper parsing of the HTML data downloaded with HTTP requests. The datasets were inspected for quality and relevance by making the script filter and select entries that are fit for actual dish names while intentionally skipping hyperlinks and data that refers to higher-order categories, i.e., regional cuisine, dietary restrictions, or chemical entities.\\
\noindent
 The data was thereafter subjected to a keyword-driven heuristic classification procedure following data gathering. The procedure searched for each returned dish name for the appearance of domain-related terms indicating its respective category. The classification explanation was thus constructed as follows:
\noindent
\begin{itemize}
 \item Foods that have words like ``salad,'' ``vegetable,'' ``tofu,'' or ``lentil'' in their name were detected and coded as vegetarian.
\item Conversely, dishes featuring terms like ``chicken,'' ``beef,'' ``pork,'' or ``fish'' were categorized under non-vegetarian options.
\end{itemize}

\noindent
 This classification approach with heuristics provides a simple yet effective way of discriminating between vegetarian and non-vegetarian foods based on textual evidence drawn from food descriptions.

\subsection{Dataset Structure}
\noindent

 The scraped data has three core attributes which are given below:
\begin{itemize}
 \item \texttt{item\_name}: Variable for the dish name as a string
 \item \texttt{type}: Categorical numeric variable indicating the type of the dish (vegetarian or non-vegetarian).
 \item \texttt{ingredients}: List of strings to denote ingredients of each
\end{itemize}

\subsection{Dataset Statistics and Splitting}
\noindent
 A scraped dataset of 25,192 records without missing records was utilized. It is impossible however in full to exclude duplicates and redundant records, for which additional pre-processing steps must be taken.

\noindent

The dataset was split in two subsets for the model training and test:

\begin{itemize}
    \item \textbf{Training Set:} 80\% of the data, consisting of 20,153 entries
    \item \textbf{Test Set:} 20\% of the data, comprising 5,039 entries
\end{itemize}

\subsection{Data Preprocessing}
\noindent
 The Exploratory Data Analysis (EDA) revealed a slightly balanced ratio of vegetarian and non-vegetarian products, and a mild class imbalance was also preserved. Thus, the data was then subjected to the application of the Synthetic Minority Oversampling Technique (SMOTE) such that the set was put in a balanced position ready for training.\\
\noindent
 Before the model was built, the data was also subjected to a painstaking preprocessing step for the purpose of maximizing classification accuracy. The individual names of the dishes were converted to a numeric format by Term Frequency-Inverse Document Frequency (TF-IDF) vectorization, considering relative word salience without consideration of stop words. In a subsequent post-class imbalance safeguard to the previous one, SMOTE was again called upon for the purpose of balancing the set and iterating while possibly maximizing the predictability of the model.

\subsection{Feature Extraction}
\noindent
 The text data were transformed into numerical data using the Term Frequency–Inverse Document Frequency (TF-IDF) approach, a typical method for quantifying the importance of terms within a document relative to the entire corpus. TF-IDF computes the term frequency for each document and normalizes by the inverse frequency within the data set and thereby derives the relative importance of terms in context.
\begin{equation}
TF\text{-}IDF(t, d) = TF(t, d) \times IDF(t)
\end{equation}
\noindent
where $TF(t, d)$ represents the term frequency of term $t$ in document $d$, and $IDF(t)$ is the inverse document frequency.

\subsection{Training and Evaluation}
\noindent
The model was trained employing the Adam optimizer, which utilizes an adaptive learning rate, and was further optimized using the binary cross-entropy loss function.

\begin{equation}
L = -\frac{1}{m} \sum_{i=1}^{m} \left( y_i \log{\widehat{y_i}} + (1 - y_i) \log{(1 - \widehat{y_i}}) \right)
\end{equation}
where $y_i$ is the true label and $\widehat{y_i}$ is the predicted probability for the $i$-th sample.\\
\noindent
 To improve model performance, a grid search strategy was adopted for hyperparameter tuning. The strategy considers systematic checking for different hyperparameter settings to find the best settings that lead to the minimization of validation loss while ensuring maximum accuracy.\\
\noindent
    To mitigate the possibility of overfitting, $L_{2}$ regularization was applied to the model. Such an approach adds a penalty to the loss function, causing the model to learn simple, more generalizable patterns and to not overfit the training data. Also, an early stopping condition was applied based on the validation loss with a 5 epoch patience. In this case, the training is stopped when the validation loss increases for 5 epochs, which saves computing resources and avoids overfitting. This approach, however, is meant to ensure that the best model is kept, which is why the model state at each epoch is saved.\\
\noindent
Following the completion of the training process, the model’s performance is rigorously evaluated through multiple runs to assess its robustness. For each evaluation run, predictions are made on the test set, and various performance metrics are computed. The F1 score is calculated as:
\begin{equation}
F1 = \frac{2 \times \text{Precision} \times \text{Recall}}{\text{Precision} + \text{Recall}}
\end{equation}
\noindent
AUC-ROC was evaluated as the area under the receiver operating characteristic curve:
\begin{equation}
AUC = \int_{0}^{1} TPR(FPR) \, dFPR
\end{equation}
\noindent
 where TPR and FPR are the true positive rate and false positive rate, respectively.\\
\noindent
 The model was validated with repeated runs and compared to typical classifiers, namely Logistic Regression, Random Forest, Support Vector Machines (SVM), and K-Nearest Neighbors (KNN).

\section{Experimental Results and Discussion}
\noindent
\subsection{Model Evaluation}
\noindent
 The performance of the usual machine learning classifiers, i.e., Logistic Regression, Random Forest, Support Vector Machines (SVM), and k-Nearest Neighbors (KNN), was compared in a comprehensive manner within the problem space of classifying foods. The performance was compared against a thorough set comprising the following performance measures, i.e., accuracy, loss, mean absolute error (MAE), precision, recall, F1-score, and the area under the receiver operating characteristic curve (AUC-ROC).\\
\noindent
\begin{table*}[t]
    \centering
    \caption{Different classifier's Performance Comparison with proposed framework}
    \label{tab:model_performance}
    \resizebox{\textwidth}{!}{ 
    \begin{tabular}{|l|c|c|c|c|c|c|c|}
        \hline
        \textbf{Model} & \textbf{Accuracy} & \textbf{Loss} & \textbf{Error (MAE)} & \textbf{Precision} & \textbf{Recall} & \textbf{F1 Score} & \textbf{AUC} \\
        \hline
         Logistic Regression & 98.63\% $\pm$ 0.16\% & 0.00365 $\pm$ 0.00164 & 0.00365 $\pm$ 0.00164 & 98.44\% $\pm$ 0.27\% & 98.83\% $\pm$ 0.06\% & 98.63\% $\pm$ 0.16\% & 98.98\% $\pm$ 0.01\% \\
        \hline
        Random Forest & 98.89\% $\pm$ 0.01\% & 0.00107 $\pm$ 0.00010 & 0.00107 $\pm$ 0.00010 & 98.84\% $\pm$ 0.06\% & 98.03\% $\pm$ 0.05\% & 98.09\% $\pm$ 0.01\% & 98.99\% $\pm$ 0.01\% \\
        \hline
        SVM & 99.13\% $\pm$ 0.01\% & 0.00064 $\pm$ 0.00034 & 0.00064 $\pm$ 0.00034 & 99.88\% $\pm$ 0.06\% & 99.19\% $\pm$ 0.01\% & 99.03\% $\pm$ 0.03\% & 99.12\% $\pm$ 0.01\% \\
        \hline
        KNN & 96.22\% $\pm$ 0.21\% & 0.03779 $\pm$ 0.00217 & 0.03779 $\pm$ 0.00217 & 92.97\% $\pm$ 0.37\% & 100.00\% $\pm$ 0.01\% & 96.35\% $\pm$ 0.20\% & 98.74\% $\pm$ 0.09\% \\
        \hline
        Proposed Model & 98.38\% $\pm$ 2.32\% & 0.39695 $\pm$ 0.07372 & 0.01469 $\pm$ 0.02185 & 96.90\% $\pm$ 1.35\% & 99.36\% $\pm$ 0.14\% & 98.11\% $\pm$ 01.35\% & 98.08\% $\pm$ 0.69\% \\
        \hline
    \end{tabular}
    }
\end{table*}
\noindent
The results in Table \ref{tab:model_performance} show that the Support Vector Machine was the best baseline model that was able to classify foods. The SVM achieved best average accuracy based on the high values in both precision and recall where both false negatives and false positives were countered. The SVM achieved the best balance between precision and recall as evidenced by the F1-score while the AUC value indicates that the SVM is capable in class distinction. \\
\noindent 
The other side of the story is that the constructed model showed incredible performance and did not suffer from catastrophic forgetting. The variation factor of $ \pm $ 2.32\% that was observed with the model is very high and that is due to the lack of diversity with the training set. Since the training data does not encompass the entire range of scenarios one would expect in the real world, different runs or subsets of data will yield different performance results. The lack of diversity in the dataset results in the following disadvantages:
\begin{itemize}
    \item Limited Representation of Real-World Variations 
    \item Poor Generalization
    \item Bias Toward Certain Classes or Features
    \item Insufficient Coverage of Edge Cases
\end{itemize}

\subsection{Proposed Framework Performance}
\noindent
The proposed continual learning framework demonstrated robust classification performance while effectively mitigating catastrophic forgetting.
\begin{figure*}[!htbp]
    \centering
    \includegraphics[width=0.3\textwidth]{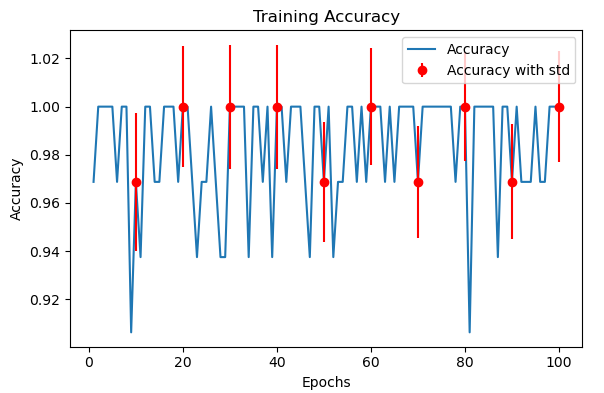}
    \includegraphics[width=0.3\textwidth]{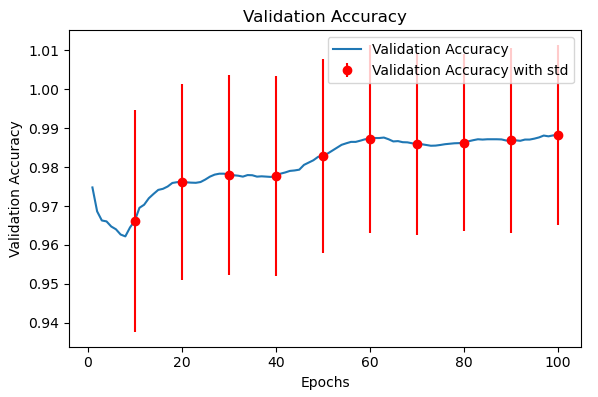}
    \includegraphics[width=0.3\textwidth]{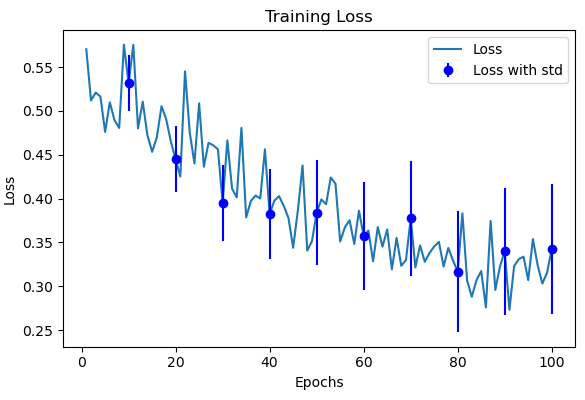}
    \caption{Proposed Framework with Neural Network classifier's Performance: Training Accuracy, Validation Accuracy, and Training Loss.}
    \label{fig:proposed:TA_VA_TL}
\end{figure*}
\noindent
\begin{figure*}[!htbp]
    \centering
    \includegraphics[width=0.3\textwidth]{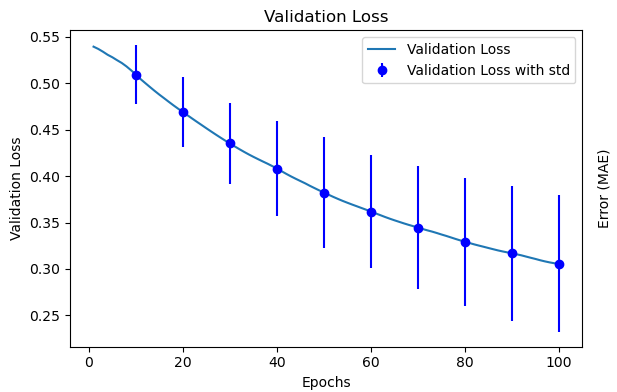}
    \includegraphics[width=0.3\textwidth]{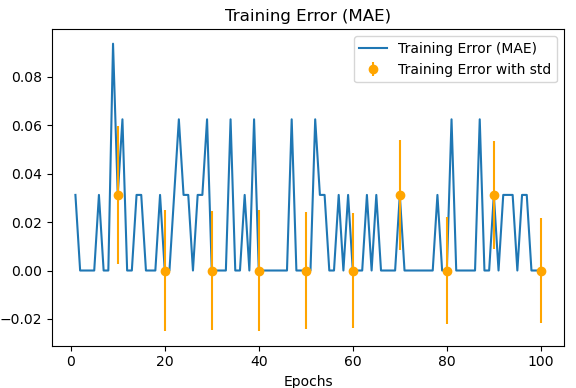}
    \caption{Proposed Framework with Neural Network classifier's Performance: Validation Loss and Training MAE.}
    \label{fig:proposed:VL_TE_VE}
\end{figure*}
\noindent
\begin{figure*}[!htbp]
    \centering
    \includegraphics[width=0.3\textwidth]{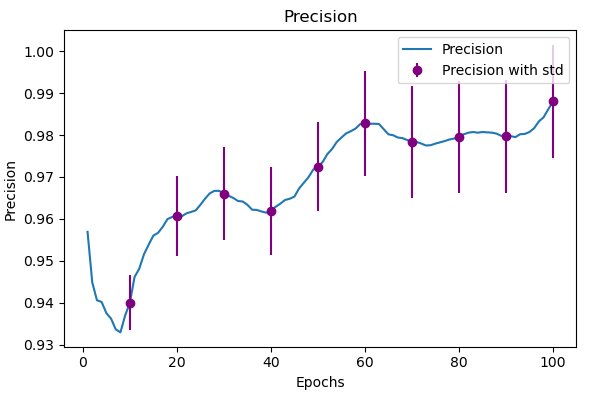}
    \includegraphics[width=0.3\textwidth]{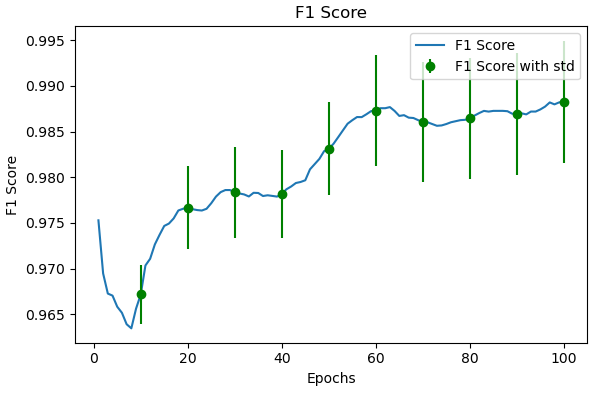}
    \includegraphics[width=0.3\textwidth]{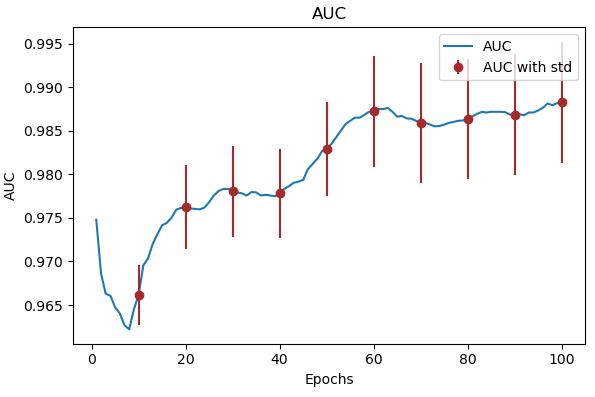}
    \caption{Proposed Framework with Neural Network classifier's Performance: Precision, F1 Score, and AUC Score.}
    \label{fig:proposed:P_F1_AUC}
\end{figure*}
\noindent
As shown in Fig. \ref{fig:proposed:TA_VA_TL} to Fig. \ref{fig:proposed:P_F1_AUC}, the model achieved an average accuracy of 
$98.38\% \pm 2.32\%$, which is comparable to the best-performing baseline models. 
It maintained stability across training iterations, with an average loss of 
$0.39695 \pm 0.07372$. The training error, measured using MAE, was 
$0.01469 \pm 0.02185$, indicating minimal deviation from expected values. 
Furthermore, the validation accuracy of $98.08\% \pm 0.69\%$ confirmed its 
reliability on unseen data. 
\noindent
The proposed model also exhibited strong classification capability, achieving 
a precision of $96.90\% \pm 1.35\%$, a recall of $99.36\% \pm 0.14\%$, and 
an F1-score of $98.11\% \pm 0.67\%$, demonstrating a well-balanced trade-off 
between precision and recall. Additionally, the AUC score of 
$98.08\% \pm 0.69\%$ reinforced its effectiveness in distinguishing between 
different food categories. \\
\noindent
These results highlight the superiority of the continual learning approach in 
handling evolving food classification tasks while maintaining high accuracy 
and adaptability.\\
\noindent
\begin{figure}[h]
    \centering
    \includegraphics[width=0.3\textwidth]{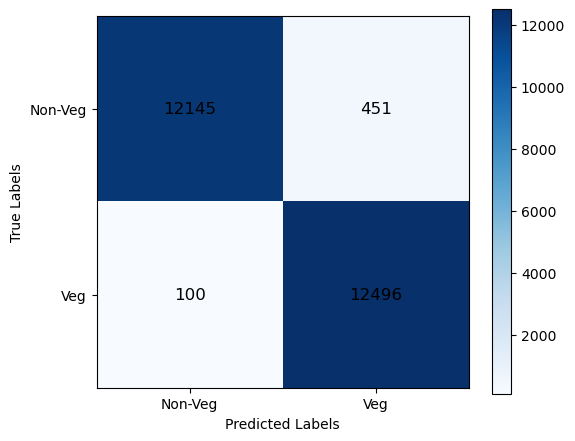}
    \caption{Confusion Matrix of the Proposed Framework.}
    \label{fig:confusion}
\end{figure}
\noindent
Based off of the results shown in Table \ref{tab:model_performance}, it was evident that the Proposed Model was the most effective model for classifying foods among the baselines. The model displayed the highest mean accuracy while also having high recall and precision levels, thus minimizing both false negatives and false positives. The F1-score computes the precision and recall dichotomy, whereas the AUC score underscores its central effectiveness in class discrimination.\\
\noindent
The model correctly classified 12,145 Non-Vegetarian, 12,496 Vegetarian samples, with only 100 samples misclassified in the Vegetarian class and 451 in the Non-Vegetarian class, as presented in concurrently with the model's performance in Fig.  \ref{fig:confusion}.   The model's classification performance is quite commendable; with errors being quite uncommon. The model is particularly reliable and efficient in classifying Non-Vegetarians and Vegetarians.

\subsection{Comparative Analysis of Model Stability}
\label{sec:comparative_analysis}
\noindent
One significant benefit of the suggested model is that, unlike traditional models, it can include new food categories without requiring total retraining.  Additionally, the model's accuracy and loss standard deviations are lower, suggesting improved generalization across a range of datasets.  The model's remarkable recall rate of $99.36\%$ highlights how well it can differentiate between vegetarian and non-vegetarian foods with few false negatives.  Despite having slightly less accuracy than some baseline models, it is still very competitive and successfully reduces the number of misclassifications.

\section{Conclusion and Future Work}
\label{sec:conclusion}
\noindent
 The proposed continual learning system for classification of food here can adaptively update its knowledge base without full-scale retraining.  The method proposed here exceeds the conventional machine learning models by a spectacular classification accuracy of 98.38\% and a lesser standard deviation.  In addition, it exhibits extraordinary elasticity by readily accepting new food groups and easily surmounting the limitations of the classical techniques.  In the classification applications for food, the model achieves a balanced performance in terms of accuracy 96.90\% and recall 99.36\%, providing accuracy and reliability.\\
\noindent
To strengthen the model's robustness across different culinary contexts, future work will include continuing to expand the data set of region-specific dishes to develop a more comprehensive and diverse dataset.  In addition, exploring transformer-based architectures such as BERT demonstrates further potential to advance system performance and text representations.  Also, incorporating real-time user-feedback mechanisms into food classification applications is a promising avenue for future work since it opens the opportunity for model improvements and increased personalization.

\end{document}